\documentclass[letterpaper, 10 pt, conference]{ieeeconf}  
\pdfoutput=1
\usepackage[dvips]{graphicx}

\IEEEoverridecommandlockouts                              

\usepackage{multicol}
\usepackage{mathtools}
\usepackage{amsmath}
 \numberwithin{equation}{subsection}
\usepackage{graphics} 
 \usepackage{amsmath}
 \usepackage{tabularx}
 \usepackage{graphicx}
 \usepackage{cite}
\usepackage{sidecap}  

\usepackage[font=small,skip=0pt]{caption}

\usepackage[table]{xcolor}
\usepackage{algorithm}
\usepackage{algpseudocode}
\usepackage{placeins}
 \usepackage{url}
\makeatletter
\def\BState{\State\hskip-\ALG@thistlm}
\makeatother
\title{\LARGE \bf Evaluation of Adversarial Training on Different Types of Neural Networks in Deep Learning-based IDSs
}

\author{ \parbox{4 in}{\centering Rana Abou Khamis and Ashraf Matrawy
       \thanks{}\\
       School of Information Technology\\
       Carleton University, Ottawa, Ontario, Canada.\\
       {\tt\small \{rana.aboukhamis,ashraf.matrawy\}@carleton.ca}
        }
}
\begin{document}

\maketitle
\thispagestyle{empty}
\pagestyle{empty}

\begin{abstract}
Network security applications, including intrusion detection systems of deep neural networks, are increasing rapidly to make detection task of anomaly activities more accurate and robust. With the rapid increase of using DNN and the volume of data traveling through systems, different growing types of adversarial attacks to defeat them create a severe challenge. In this paper, we focus on investigating the effectiveness of different evasion attacks and how to train a resilience deep learning-based IDS using different Neural networks, e.g., convolutional neural networks (CNN) and recurrent neural networks (RNN). We use the min-max approach to formulate the problem of training robust IDS against adversarial examples using two benchmark datasets. Our experiments on different deep learning algorithms and different benchmark datasets demonstrate that defense using an adversarial training-based min-max approach improves the robustness against the five well-known adversarial attack methods.

\end{abstract}


\section{Introduction}

DNN are increasingly used in security applications such as network intrusion detection systems (IDS) \cite{ibitoye2019analyzing}\cite{abou2019investigating}. The fact that IDS is adversarial in nature, makes security aspects of DNN increasingly important and a critical design goal, especially against adversarial samples. 
While deep learning-based IDS aim to be very effective in classification between benign and malign inputs, adversarial samples often expose blind spots in the inputs. Researches on adversarial machine learning concentrate on two main points: how to generate capable adversarial samples that can deceive a model with small perturbation and how to build and defense a model to be robust against adversarial samples. The protection and robustness against adversarial attacks is a growing challenge and should always be addressed by DNN researchers and designers. 
Several works have studied adversarial attacks for DNN in image classification and after reviewing the literature in the field of adversarial attack against deep learning-based IDS \cite{ibitoye2019analyzing}\cite{abou2019investigating}, to the best of our knowledge, it is still the early stages and there is a lack in the current studies demonstrated optimization and robustness of deep learning-based IDS.
This served as a motivation to explore in-depth deep learning-based IDS and focus on different algorithms of deep neural networks, including CNN and RNN, on different benchmark datasets and study the effectiveness of different adversarial attack methods against DNN and their robustness. In this work, we use min-max approach  \cite{madry2017towards} \cite{al2018adversarial} as extension of\cite{abou2019investigating}. AbouKhamis et. al \cite{abou2019investigating} applied min-max in deep learning-based IDS focusing on one type of deep neural network: Feed-Forward Neural network using one benchmark dataset and three adversarial attack methods based on gradient method: FGSM, Bit Gradient Ascent (BGA) and Bit Coordinate Ascent(BCA)\cite{al2018adversarial}.
While in this work, our main contributions are demonstrating that the min-max problem is manifestly beneficial in different type of DNN. This leads to focusing on the following:

\begin{itemize}
\item We work on both sides: attack and defense IDS-based deep learning using different DNN algorithms: ANN, CNN, and RNN by utilizing resilient deep learning-based IDS framework based on min-max approach \cite{abou2019investigating} to generate a powerful adversarial sample and augment them during training time. 

\item We use five well-known adversarial attacks, including a simple one-step attack such as Fast Gradient Sign Method (FGSM), to more powerful attacks with multi-step like BIM, Projected Gradient Descent (PGD), Carlini and Wagner (CW) and Deepfool. Those attacks are used to solve the inner maximization problem in the min-max formulation.

\item Build six adversarial-free deep learning-based IDS with three different architecture: ANN, CNN, and RNN on two benchmark datasets: UNSW-NB 15 and NSD-KDD as baseline IDS models using two pre-processing methods: Recursive  Feature  Elimination (RFE) and Principal Component Analysis (PCA).

\item We retrained thirty adversarial IDS models to investigate the min-max approach in more detail. 
\end{itemize}

We structure the remainder of the paper as follows. In section II, we provide a background of adversarial samples and thread model. In Section III, we present a literature review of some related work in deep learning-based Intrusion detection. In Section IV, We explorer our methodology including the IDS framework and algorithm. The experimental results and evaluation are given in Section V and VI. Section VII briefly summarize this paper.

\section{Adversarial Threat Model and Attacks}
In this work, we have a classification problem where the DNN algorithm goal is to learn the Decision boundary to be able to classify input data into two classes: attack (positive) or benign (negative).
\subsection{Adversarial Threat Model}

\begin{figure}[htbp]
\includegraphics[width=\linewidth,keepaspectratio=true]{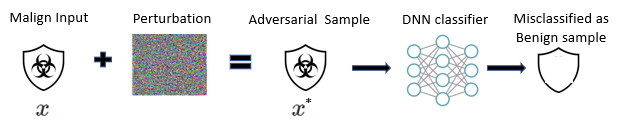}
  \caption{Generating adversarial samples by adding small perturbation to the original malign packet to misclassified as benign packet}
  \label{fig:ad}
  \centering
  \vspace{-4.0mm}
\end{figure}

The fact that the training of DNN is based on data, classification tasks can be carefully manipulated by crafted, and perturbations inputs called adversarial samples that are often visually imperceptible to mislead a model toward incorrect classification and evade detection reliably. Adversarial attacks can attempt to attack data or models in a way to make the decision boundary between the regular and the crafted data shift and inaccurate.
Steps for adding noise to the original sample are illustrated in Fig \ref{fig:ad}.
By using inner-maximizer approach \cite{abou2019investigating}, the adversarial attack can defeat classifiers by adding a calculated perturbation $\gamma$ to legitimate samples $x$ to create a new version $x^*$ called "Adversarial Sample" that maximize the loss


To harden the IDS models, the proposed IDS framework \cite{abou2019investigating} incorporates the adversarial samples generated by the maximization formula in \cite{abou2019investigating} during training time. This can be done by combine minimization formula with maximization formula in the min-max formulation as done in\cite{abou2019investigating}\cite{al2018adversarial}.

We evaluated the studied min-max methods on UNSW-NB 15 and NSL-KDD datasets of a packet flow, where the task is to classify packet flow into the correct class. In this work, we consider some assumptions about our adversarial threat model. The attacks are evasion attack \cite{ibitoye2019threat}, where an attacker has access to the IDS models during prediction time that leads to misclassify the model decision and target the positive "attack" sample to be classified as a negative "benign" sample taking into considerations that a complete knowledge of the targeted models is known to perform a white-box attack with multi-iterations.


\vspace{-3.0mm}

\vspace{-2.0mm}
\subsection{Adversarial Attack Methods}
We use five methods to generate adversarial samples. We use Fast Gradient Sign Method (FGSM) \cite{goodfellow2014explaining} with a step gradient update with the direction of the gradient of the loss function. Basic Iteration method (BIM), which runs multiple iterations of FGSM  and avoid big changes in input features \cite{kurakin2016adversarial}, is the second method. In the third method, we implement a Projected Gradient Descent attack (PGD), which also attempts to find the perturbation that maximizes the loss taking into consideration keep the epsilon perturbation small enough to lies in the permitted range \cite{madry2017towards}. We also implement one of the most effective white-box attacks called Carlini and Wagner (CW) \cite{carlini2019evaluating} that defeat defensive distillation and considered to be used to evaluate the resilience of the potential models. The fifth attack is Deepfool \cite{moosavi2016deepfool} attack that is designed based on the concept of generalizable, which means that adversarial samples generated for a particular model can also fool other models. All listed adversarial attacks were initially designed and tested on image classification models  \cite{al2018adversarial} \cite{shaham2018understanding}.
\vspace{-3.0mm}
\section{Related Work}
\vspace{-1.0mm}
In this section, we highlight the related work for adversarial machine learning with a combination of adversarial attack methods and defense techniques. In particular, adversarial attacks in deep learning-based IDS, as shown in Table \ref{tab:IDS_Related_work}.

Some working including \cite{al2018adversarial} \cite{madry2017towards} \cite{abou2019investigating} \cite{shaham2018understanding} utilizing min-max formulation and deep learning techniques to design optimal intrusion detection system that can robustly handle different types of adversarial attacks. In this paper, we built an experimental prototype based on \cite{abou2019investigating} to
incorporate generated adversarial samples during the training process and successfully achieve
robustness against the adversarial samples.

\begin{table*}[t]
\caption{Machine Learning-based IDS Related Work}
\label{tab:IDS_Related_work}
\resizebox{\textwidth}{!}{%
\begin{tabular}{  |l|c|l|l|l|l|} 
\hline
 Paper & Year & Algorithms & Datasets & Adversarial Attacks & Defense \\
    \hline\hline
    Aboukhamis et al.\cite{abou2019investigating} & 2019 & ANN & UNSW-NB 15 
 & FGSM, BGA, BCA & min-max approach  \\
    \hline
   Ibitoye et al.\cite{ibitoye2019analyzing} & 2019 & FNN,SNN & BoT-IoT 
 & FGSM, BIM, PGD &  Feature Normalization  \\
    \hline
    YePeng et al. \cite{peng2019evaluating} & 2019 & DNN, SVM, RF,LR &NSL-KDD&
  FGSM,PGD,L-BFGS,SPSA & N/A\\
  \hline
  Biggio et al. \cite{biggio2013security} & 2013& SVM & HTTP-delivered attacks & Causative (Poisoning) Attack & N/A \\
  \hline
  Wang et al. \cite{wang2018deep} & 2018 & DNN & NSL-KDD &FGSM,JSMA,Deep fool,CW & N/A\\
 
\hline
  Biggio et al. \cite{biggio2011bagging}&2011&SpamBayes spam Filter model & Lab real traffic& Causative (Poisoning) Attack& Bagging Ensembles\\
 \hline
  Homoliak et al. \cite{homoliak2018improving}& 2018 & N Bayes,Log. Reg., D. Tree, SVM & ASNM-NPBO & Evasion Attack &   obfuscations-aware classiﬁer \\
 \hline
 Kloft et al. \cite{kloft2010online}&2010&  Support vector data description (SVDD) & real HTTP traffic& Causative (Poisoning) Attack & N/A\\
 \hline
 Javaid et al. \cite{javaid2016deep}&2016&  DNN & NSL-KDD& N/A& N/A\\

 \hline
Kwon et al. \cite{kwon2017survey}&2017& DNN Survey & KDDCup 1999 & N/A& N/A\\
\hline
 Shone et al. \cite{shone2018deep} &2018 &DNN & NSL-KDD & N/A& N/A\\
 \hline
Yin  et al. \cite{yin2017deep} &2017 &RNN & NSL-KDD & N/A& N/A\\
  \hline
Khalid  et al. \cite{alrawashdeh2016toward} & 2016 & Restricted Boltzmann Machine(RBM), Logistic Regression (LR)& KDDCup 1999 & N/A& N/A\\
 
 \hline
\end{tabular}}
\end{table*}

\vspace{-2.0mm}
\section{Methodology and Experimental Approach}
We first train the DNN models in two different phases: Phase I and Phase II. 
In Phase I, we first split the two benchmark datasets: UNSW-NB 15 and NSD-KDD into training, testing, and validation sets. We then pre-process the datasets and apply feature selection using two techniques: PCA and Recursive Feature Elimination (RFE). Experiment results of applying two feature selection techniques determine which feature selection method to move forward with the entire experiments. We then train six baseline deep learning-based IDSs: ANN, CNN, and RNN on two adversarial-free datasets to attack them in Phase II by five different types of adversarial samples.

In Phase II, we have conducted various sets of experiments to generate different adversarial samples: FGSM, BIM, PGD, CW, and Deepfool using inner-maximizer and evaluate their effectiveness against the baseline models. In the second part of Phase II, we evaluate the robustness of the deep learning-based IDS framework architectures: ANN, CNN, and RNN that are retrained using \textbf{\emph{Algorithm I}}.





The adversarial training process based on the min-max approach is formalized in \textbf{\emph{Algorithm I}}. The algorithm describes the IDS framework solution based on min-max formulation \cite{abou2019investigating}.
The deep learning-based IDS framework solution is illustrated in Fig \ref{fig:dnn}. 

All our models were implemented using TensorFlow and Keras. The experiments were performed on Virtual machine with GPU acceleration with 32 GB memory and three Intel core processors 2.294 GHz. For generating adversarial samples, we use the open-source IBM Robustness Toolbox (ART) framework \cite{nicolae2018adversarial}.
\vspace{-1.0mm}
\subsection{Data sets and Pre-processing }
In this paper, we use two datasets: UNSW-NB15 \cite{moustafa2015unsw} and NSL-KDD \cite{tavallaee2009detailed}.
We perform different methods for pre-processing.
Both datasets have different scales and some significant outliers. To enhance training time and remove outliers, we re-scale all feature values into a range between 0 and 1.  We use an encoder to convert the categorical data into numbers. 
\subsubsection{UNSW-NB 15 Dataset}
It contains nine different attacks, includes DoS, worms, Backdoors, and Fuzzers. In our experiments, we use binary classification with two classes: benign and attack labels, and leave the multi-class for further investigations. Each row in the dataset has 49 features, including class labels, and it has more than two million records.
We use in our experiments 1,17478 records for training and 57,863 for testing. 

\subsubsection{NSL-KDD Dataset}
It is a new version of the KDD 99 dataset that solves various problems of the KDD 99 dataset. It has 21 attack types in the training dataset and 37 attacks. We use binary classification with two classes: benign and attack label.
Our training set consists of approximately 100,778 flow and 25,195 to the testing set.
  
\vspace{-1.0mm}

\begin{algorithm}[t]
\label{algo}
\caption{Experimental Process} 
\begin{algorithmic}[I]
\State\textbf{Input: }{ Training set, M:inner maximizer}
\State \textbf{Output: }{Robust Adversarial trained model, $x^*$ adv. samples}
\State Load Train, Validation, Test sets
\State Extract Important features
\State Construct ANN, CNN and RNN model C
\State Define inner-maximization M
\State $Batch \leftarrow 10$
\Repeat 
\State Read $Batch$ of 32 samples 
\If {Evasion method \textbf{!=} \emph{Attack-free}}
\State Generate Adv Samples using M($Batch^*$)
\State Start Adversarial Learning ($Batch^*$)
\State do Test($Batch^*$)
\EndIf
\Until{epoch=10 and C network converged}

\end{algorithmic}
\vspace{-1.0mm}
\end{algorithm}

\begin{figure*}[htbp]
 \centering
 \includegraphics[width=0.7\textwidth,keepaspectratio=true]{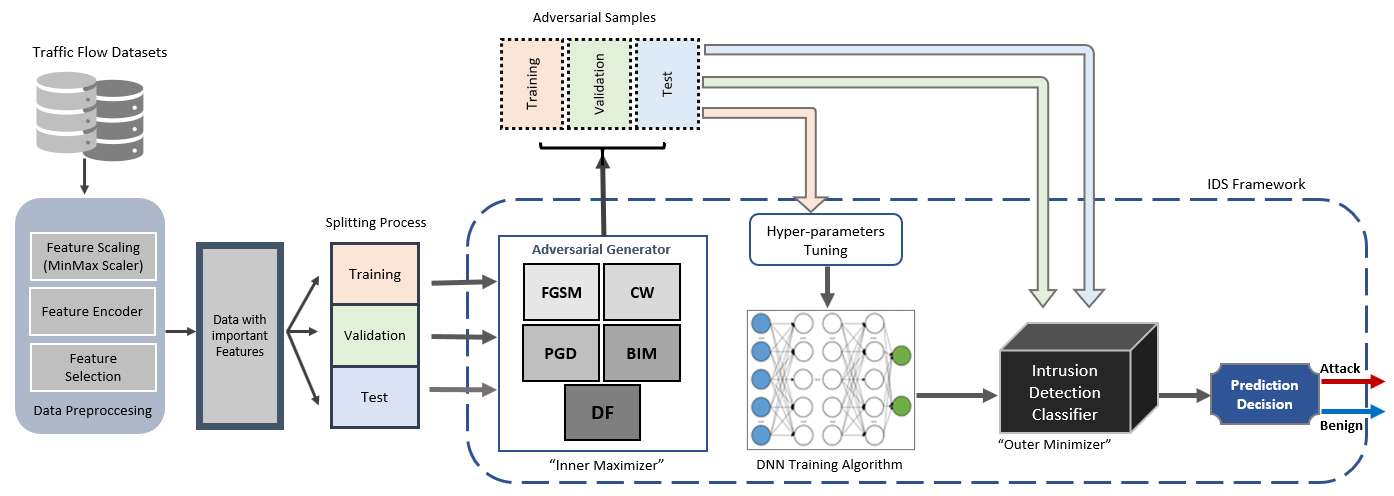}
 \caption{IDS Framework Architecture with the inner-maximizer that generate adversarial samples and the outer-minimizer that increase robustness adapted from \cite{abou2019investigating}}
  \label{fig:dnn}
  \vspace{-4.0mm}
\end{figure*}

In all experiments, we tested the adversarially trained model using new adversarial samples that are not used in the training time. Using different samples in a testing grantee that the model evaluated using unseen samples. 

\subsection{Architecture Characteristics and Learning Setup}
ANN architecture and hyper-parameters listed in Table \ref{tab:ann_para}. 
CNN model built with same architecture and hyper-parameters for both UNSW and NSL-KDD as listed in Table \ref{tab:cnn_para}. Table \ref{tab:rnn_para} has the RNN hyper-parameters.

\begin{table}[h]
\caption{Training Parameters and Architecture  for IDS Models in ANN Experiments}
 
\label{tab:ann_para}
\begin{center}
\begin{tabular}{ |l|l| } 
\hline
Parameter & Value\\ 
\hline
\hline
No. of hidden layers & 3\\
\hline
Layer 1 & 128 neurons\\
\hline
Layer 2 & 96 neurons\\
\hline
Layer 3 & 64 neurons\\
\hline
Dropout & 0.25\\
\hline
Optimizer & ADAM\\
\hline
Activation function & ReLU and Softmax\\
\hline
Learning rate & 0.01\\
\hline
Epoch & 10\\
\hline
Batch Size & 32 \\
\hline
\end{tabular}
\end{center}
\vspace{-5.0mm}
\end{table}

\begin{table}[h]
\caption{Training Parameters and Architecture  for IDS Models in CNN Experiments}
\label{tab:cnn_para}
\begin{center}
\begin{tabular}{ |l|l| } 
\hline
Parameter & Value\\ 
\hline
\hline
Convolution layer  & 3 layers with ReLU\\
\hline
Max Pooling & 2 layers\\
\hline
Dropout & 0.25\\
\hline
Fully Connected & 4 layers with ReLU\\
\hline
Output layer& Softmax\\
\hline
Optimizer & ADAM   \\
\hline
Activation function & ReLU and Softmax  \\
\hline
Learning rate & 0.01\\
\hline
Epoch & 10 \\
\hline
Batch Size & 32 \\
\hline
\end{tabular}
\end{center}
\vspace{-5.0mm}
\end{table}

\begin{table}[th!]
\caption{Training Parameters and Architecture  for IDS Models in RNN Experiments}
\label{tab:rnn_para}
\begin{center}

\begin{tabular}{ |l|l| } 
\hline
Parameter & Value\\ 
\hline
\hline
LSTM layer  & 2 layers with Sigmoid\\
\hline
Fully Connected & 1 layer with ReLU\\
\hline
Dropout & 0.5\\
\hline
Output layer& Softmax\\
\hline
Optimizer & ADAM   \\
\hline
Activation function & ReLU , Softmax and Sigmoid \\
\hline
Learning rate & 0.01\\
\hline
Epoch & 10 \\
\hline
Batch Size & 32 \\
\hline
\end{tabular}
\end{center}
\vspace{-5.0mm}
\end{table}
\vspace{-2.0mm}
\subsection{Evaluation Metrics}
To understand the robustness of trained models in the adversarial environment, we use different metrics to evaluate the performance. 
Prediction Accuracy (AC): Total number of correctly classified samples from benign and attack samples among all number of all samples. Specificity (or Precision): It is a number of true positives among samples classified as positive. Sensitivity( or Recall): It is called True Positive Rate (TPR), it is a number of True Positives (TP) among all positive samples.
To evaluate the overall performance of our binary models among several algorithms, ANN, CNN, RNN,  we utilized a useful metrics called Area Under the Curve (AUC). It is a number that reflects the model performance. The best model will have a higher AUC toward 1. 
\vspace{-3.0mm}

\section{Phase I: Performance of adversarial-free Deep learning-based IDS }
Our objective is to build the most accurate deep learning model on UNSW-NB and NSD-KDD for each architecture: ANN, CNN, and RNN as baseline models (benchmarks) to comprehensively compare the performance of Deep learning-based IDSs in the adversarial environment in Phases II. 
We build six baseline IDSs: ANN-UNSW, ANN-KDD, CNN-UNSW, CNN-KDD, RNN-UNSW, and RNN-KDD. 
\vspace{-2.0mm}
\subsection{Feature Selection}
In this section, We focus on investigating experimentally the best feature selection methods that can be applied to achieve high accuracy and better performance of deep learning-based IDSs. We use two different selection techniques, Recursive Feature Elimination (RFE) and Principal.


\subsubsection{Performance using RFE}
We train the ANN, CNN, and RNN using RFE feature selection for UNSW and NSD-KDD. We first selected the top seven features. 
The experiment was executed multiple times for each DNN algorithms with different batches and epoch values to obtain the best accuracy. The accuracy of ANN, CNN, and RNN models on UNSW-NB and NSL KDD using RFE achieve 88.77\% as an average using the top seven features.

\begin{figure}[htbp]
    \centerline{
     \includegraphics[width=2.6in]{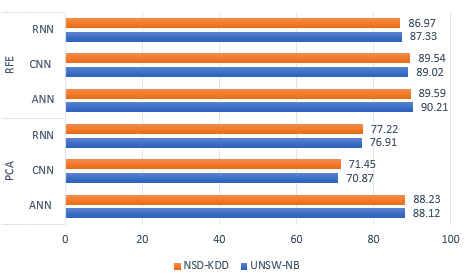}}\hfil 
    \caption{ANN, CNN and RNN IDSs accuracy for PCA and RFE}
    \vspace{-4.0mm}
    \label{fig:featuressel}
    
\end{figure}
\subsubsection{Performance using PCA}
Similarly, with the same experimental setup and parameters, we train the ANN, CNN, and RNN using PCA on UNSW-NB and NSD-KDD. We repeated the experiments three times with different number principal components 15, 10, and 5. 
The best accuracy using PCA achieved by using five principal components, as shown in Fig \ref{fig:featuressel}. 
The comparison between PCA and RFE is illustrated in Fig \ref{fig:featuressel}. We observe that accuracy with RFE for all models architectures ANN, CNN, and RNN was higher than PCA. However, we still did not achieve the efficiency that we expect. After various tries and observations, we improve our model performance by selecting the top five attributes listed in Table \ref{tab:final_features} 
This result might be possible because we use regularization techniques for prepossessing the two selected datasets.

\vspace{-2.0mm}
 \begin{table}[h]
  \centering
  \caption{The Top 5 selected features from UNSW-NB and NSD-KDD Datasets}\label{tab:final_features}
    \vspace{-3.0mm} 
  \begin{tabular}{|l|l|l|l|}
    \hline
     UNSW-NB &  NSD-KDD\\
     \hline
     \hline
     protocol&  service\\
     wrong-fragment & dbytes \\
     is-guest& rate\\
     same-srv-rate  &sload\\
     diff-srv-rate& dload\\
    \hline
  \end{tabular}
\end{table}

\vspace{-4.0mm}
\subsection{Baseline IDSs Results}
Table \ref{tab:dnn_metrics} shows the performance metrics of All baselines IDSs in adversarial-free environment using the top 5 selected features in Table \ref{tab:final_features}.
After building and evaluating the baselines deep learning-based IDS with different algorithms: ANN, CNN, and RNN on adversarial-free datasets, we are going to dive into adversarial attacks and defense experiments in the next section.
\vspace{-2.0mm}

\begin{table}[h]
\centering
\caption{Prediction Accuracy, precision, recall for baseline IDSs}
\label{tab:dnn_metrics}

\begin{tabular}{ |l|c|c|c|c| } 
\hline
Model &  Accuracy &  Precision & Recall & AUC \\ 
\hline
ANN-UNSW &    0.97 &  0.96 & 1.00 &0.99\\
ANN-KDD &    0.96 &  0.95 & 0.99 &0.98\\
\hline
CNN-UNSW &    0.96 &  0.95 & 1.00 &0.99\\
CNN-KDD &    0.96 &  0.96 & 0.96 &0.99\\
\hline
RNN-UNSW &    0.96 &  0.95 & 1.00 &0.98\\
RNN-KDD &    0.96 &  0.96 & 0.96 &0.98\\
\hline
\end{tabular}
\end{table}



\begin{figure*}[htbp]
\centering
\begin{minipage}[b]{.4\textwidth}
\includegraphics[width=\textwidth]{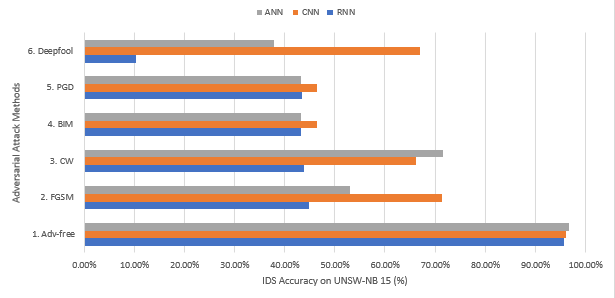}
\caption{Effect of Adversarial samples generated by inner-maximizer on ANN, CNN and RNN based IDS trained by Adversarial-free UNSW-NB}\label{fig:unswattack}
\end{minipage}\qquad
\begin{minipage}[b]{.4\textwidth}
\includegraphics[width=\textwidth]{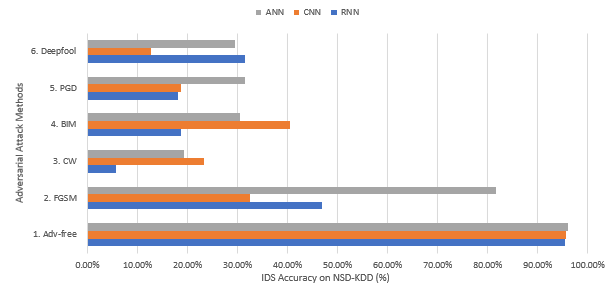}
\caption{Effect of Adversarial samples generated by inner-maximizer on ANN, CNN and RNN based IDS trained by Adversarial-free NSD-KDD}\label{fig:kddattack}
\end{minipage}
\end{figure*}

\vspace{-3.0mm}
\section{Phase II: Evaluation of the Proposed Deep Learning-based IDS Framework}
In Phase II, we report and evaluate the impact of the min-max approach. We divided Phase II into two parts: Phase II-1 and Phase II-2. 

 \begin{itemize}
 \item In Phase II-1, we start finding adversarial samples from the selected five adversarial attack methods with high perturbation using inner-maximizer.
 \item  Evaluate the quality of the adversarial samples found on UNSW-NB and NSK-KDD datasets generated by inner-maximizer against the six baseline IDS models, as shown in Fig  \ref{fig:unswattack} and Fig \ref{fig:kddattack} .
 \item In Phase II-2, we retrain the models with the same experimental setup in Phase I using Algorithm I.
 \item Evaluate the adversarially trained IDS model robustness against adversarial samples generated from the same attack method.
 \end{itemize}


 \begin{figure*}[h]
\begin{multicols}{3}
    \includegraphics[width=2.3in]{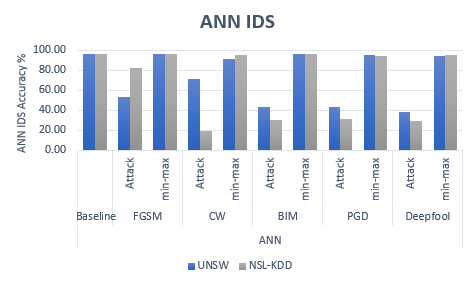}\par
    \includegraphics[width=2.3in]{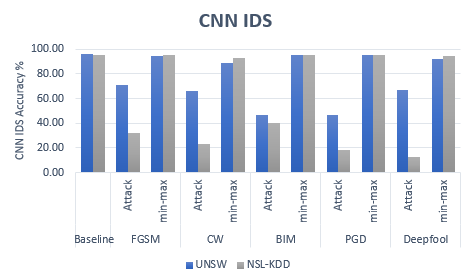}\par
    \includegraphics[width=2.3in]{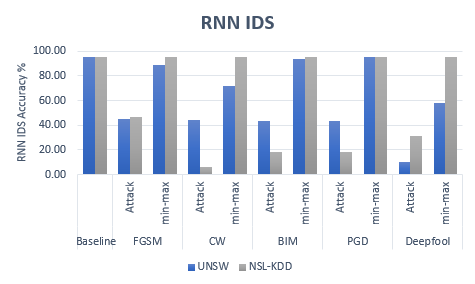}\par
    \end{multicols}
 

\caption{Prediction accuracy of ANN, CNN, and RNN before and after applying min-max on NSD-KDD and UNSW-NB 15 Datasets}

\label{fig:min-max}
\end{figure*}

\subsection{Adversarial Attacks Experiments using inner-maximizer}

The performance of ANN, CNN, and RNN based IDS in an adversarial-free environment is evident, as shown in Fig \ref{fig:unswattack} and Fig \ref{fig:kddattack}. The performance of deep learning-based IDS in all three architectures are vulnerable in the adversarial environment. Furthermore, Baseline IDS models have different robustness in UNSW-NB and NSD-KDD experiments. Fig \ref{fig:unswattack} and Fig \ref{fig:kddattack} show a significant decrease in accuracy comparing to the accuracy in adversarial-free environment.
If we look closer to ANN IDSs and evaluate their performance against all adversarial samples, we observe that BIM, PGD, and Deepfool attacks have almost the same effectiveness against ANN-UNSW and ANN-KDD. However, ANN-UNSW IDS is more resilient to CW and FGSM attacks, while ANN-KDD IDS was less resilient against CW attacks. 

If we look at CNN-UNSW IDS in Fig \ref{fig:unswattack}, we can find that BIM and PGD attacks are more destructive and decrease the accuracy by almost 50\%. While FGSM, CW, and Deepfool attacks have less effectiveness on CNN-UNSW IDS. While in CNN-KDD, CW, PGD, and Deepfool have the same and higher effectiveness comparing to BIM and FGSM attacks.

For RNN-UNSW IDS, FGSM, CW, BIM, and PGD attacks have all the same effectiveness. They decrease the initial accuracy significantly by the half. However, the Deepfool attack decreases the accuracy outstandingly by almost 85\%. CW attack against RNN-KDD outstanding other attacks. CW attacks decrease the accuracy sharply by 90\%.
Fig \ref{fig:kddattack} shows that all adversarial attacks decrease the accuracy of ANN, CNN, and RNN baseline IDS sharply.
FGSM attack has the least effectiveness among all ANN, CNN, and RNN models in NSD-KDD. 

Through the comprehensive comparison regarding the adversarial attacks against baseline IDSs, we can draw the following findings: 
FGSM Attack For ANN, CNN, and RNN models on NSL-KDD and UNSW datasets have less impact on almost all baseline models because the FGSM attack's purpose is to be fast, not optimal attack \cite{carlini2019evaluating}\cite{goodfellow2014explaining}.
We observe that BIM and PGD attacks have similar effectiveness in UNSW-NB and NSD-KDD experiments. However, BIM and PGD effectiveness against CNN-KDD are varying. 
CW attack considers the most potent attacks against baseline IDSs in the NSD-KDD experiments. Although, CW attacks one if the least effective against baseline IDSs in the UNSW-NB experiments.
FGSM, CW, BIM, and PGD attacks have relatively effectiveness in the UNSW-NB experiment.
As we observe that the DeepFool attack shows the superiority over all other attacks expects for CNN-UNSW. Deepfool attack was considered a very strong attack \cite{moosavi2016deepfool} that can generate small perturbations to evade deep learning models.

To summarise the UNSW-NB experiment, we observe that all baselines for ANN, CNN, and RNN IDS trained by adversarial-free samples are more resilient to the five adversarial attacks comparing to NSD-KDD experiments.


\subsection{Adversarial Training Experiments using min-max}
In Phase II-2, we apply \textbf{\emph{Algorithm I}} on ANN, CNN, and RNN to retrain the IDS models with the generated adversarial samples by inner-maximizer. We then evaluate the robustness of the defender adversarially trained: ANN, CNN, and RNN based IDSs on UNSW-NB and NSD-KDD with unseen adversarial samples. We also compare their performance to the baseline IDS models that we built in Phase I. 

We built ten adversarially trained models for each ANN, CNN, and RNN architectures, as shown in Fig \ref{fig:min-max}. In total, we built 30 adversarially trained IDS models on top of the six baseline IDS models. We refer to the trained models by the adversarial attack methods, as shown in Fig \ref{fig:dnn}.

Experimental results on UNSW-NB 15 and NSD-KDD datasets show a significant improvement in the prediction accuracy of the IDS models among the ANN, CNN, and RNN, as shown in Fig \ref{fig:min-max}. Figures \ref{fig:min-max} compare the prediction accuracy on UNSW-NB and NSD-KDD for all deep learning-based IDSs before an after applying the min-max approach.  

Relatively all adversarial trained models: ANN, CNN, and RNN trained using \textbf{\emph{Algorithm I}} lead to high robustness against the five types of perturbations: FGSM, CW, BIM, PGD, and Deepfool generated by inner-maximizer. The adversarially trained models are more robust to new adversarial samples and outperform the baseline IDSs.

However, we observe that the CNN-UNSW IDS trained by CW samples is less robust to new CW samples comparing to other adversarial trained models. 
Also, in RNN-UNSW IDS, we observe that Deepfool and CW models are less robust to new adversarial samples from the same attack methods comparing to other IDS models. This result corresponds to the nature of CW and Deepfool attacks. 
\cite{moosavi2016deepfool} \cite{carlini2017towards}.
FGSM, BIM, and PGD models were more robust to their adversarial samples.

\section{Conclusion and future work}

The Novelty of this research derives from the fact that it is the first experiment to implement and compare different types of architecture, e.g., ANN, CNN, and RNN, in deep learning-based IDS in an adversarial environment on two benchmark intrusion detection datasets: UNSW-NB and NSD-KDD. Most of the literature in the adversarial domain demonstrated the concept of using one deep learning architecture with different adversarial attacks. This research took the path further in investigating and evaluating the robustness of different types of DNN based IDS against five adversarial attack methods: FGSM, CW, BIM, PGD, and Deepfool using the min-max approach. We built 36 Deep learning-based IDS to show that the adversarial training based min-max approach considers a reliable defense technique against different adversarial attacks in DNN. 
For future work, the aim is to investigate if the min-max approach in deep learning-based IDS may be considered a general defense and evaluate further the robustness of adversarial trained DNN models against multiple adversarial attacks.


\section{ACKNOWLEDGEMENT}
This work was supported in part by the Natural Sciences and Engineering Research Council of Canada (NSERC) through the NSERC Discovery Grant program.
\vspace{-2.0mm}
\bibliographystyle{ieeetr}
\bibliography{references}



\end{document}